\pdfoutput=1

\documentclass[11pt]{article}

\usepackage[preprint]{acl}

\usepackage{times}
\usepackage{latexsym}

\usepackage[T1]{fontenc}

\usepackage[utf8]{inputenc}

\usepackage{microtype}

\usepackage{inconsolata}

\usepackage{graphicx}
\graphicspath{{./images/}}

\usepackage{amsmath}
\usepackage{subfigure}
\usepackage{subcaption}
\usepackage[linguistics]{forest}
\usepackage{linguex}

\usepackage{booktabs}
\usepackage{multicol}
\usepackage{mathtools}

\title{Is Structure Dependence Shaped for Efficient Communication?:\\A Case Study on Coordination}

\author{
 \textbf{Kohei Kajikawa\textsuperscript{1,2}},
 \textbf{Yusuke Kubota\textsuperscript{2}},
\textbf{Yohei Oseki\textsuperscript{1}}
\\
 \textsuperscript{1}The University of Tokyo, 
 \textsuperscript{2}NINJAL
\\
\texttt{\{kohei-kajikawa, oseki\}@g.ecc.u-tokyo.ac.jp}\\
\texttt{kubota@ninjal.ac.jp}
}

\begin{document}
\maketitle
\begin{abstract}
    Natural language exhibits various universal properties.
    But why do these universals exist?
    One explanation is that they arise from functional pressures to achieve \emph{efficient communication}, a view which attributes cross-linguistic properties to domain-general cognitive abilities.
    This hypothesis has successfully addressed some syntactic universal properties such as compositionality and Greenbergian word order universals.
    However, more abstract syntactic universals have not been explored from the perspective of efficient communication.
    Among such universals, the most notable one is \emph{structure dependence}, that is, the existence of grammar-internal operations that crucially depend on hierarchical representations.
    This property has traditionally been taken to be central to natural language and to involve domain-specific knowledge irreducible to communicative efficiency.
    
    In this paper, we challenge the conventional view by investigating whether structure dependence realizes efficient communication, focusing on coordinate structures.
    We design three types of artificial languages: (i) one with a structure-dependent reduction operation, which is similar to natural language, (ii) one without any reduction operations, and (iii) one with a linear (rather than structure-dependent) reduction operation.
    We quantify the communicative efficiency of these languages.
    The results demonstrate that the language with the structure-dependent reduction operation is significantly more communicatively efficient than the counterfactual languages.
    This suggests that the existence of structure-dependent properties can be explained from the perspective of efficient communication.

\end{abstract}

\section{Introduction}
To understand the universals of natural language, it is crucial to address \emph{why} such universals exist, as well as \emph{how} such universals can be theoretically described.
This raises the question: what kinds of pressures shape these universals?

One explanation is that the universals of natural language are shaped as a result of functional pressures to achieve \emph{efficient communication}~\citep{zipf-1949,jaeger-tily-2011,christiansen-chater-2016,kemp-etal-2018,gibson-etal-2019,futrell-hahn-2022,fedorenko-etal-2024}.
Efficient communication refers to a situation where the amount of information conveyed is maximized while the effort required for production and comprehension is minimized under human cognitive constraints.
If some structural property of languages is shaped to achieve efficient communication, it can be optimized under two competing functional pressures: the need to be as \emph{simple} as possible and the need to be as \emph{informative} as possible.

The hypothesis that two competing pressures for utilities shape the form of human language has long been assumed in linguistics~\citep{hawkins-1994,hawkins-2004,haspelmath-2008}.
In recent years, methodologies have been established to examine this hypothesis by quantifying the \emph{simplicity} and \emph{informativeness} of languages by using information-theoretic criteria~\citep{kemp-etal-2018, gibson-etal-2019, futrell-hahn-2022}.
To date, this hypothesis has been successfully examined at the lexical level~\citep[][\textit{inter alia}]{ferrer-i-cancho-sole-2003,kemp-regier-2012,piantadosi-etal-2011,piantadosi-etal-2012,regier-etal-2015,zaslavsky-etal-2018,mollica-etal-2021,steinert-threlkeld-2021,denic-etal-2022,trott-bergen-2022,uegaki-2022,chen-etal-2023,pimentel-etal-2023,van-de-pol-etal-2023,denic-szymanik-2024}.
At the syntactic level, there are pieces of empirical evidence that grammar itself is shaped to achieve efficient communication~\citep{gildea-jaeger-2015,futrell-etal-2020-lossy,futrell-etal-2020-dependency,hahn-etal-2020,hahn-etal-2021,clark-etal-2023}, and it has been shown that the existence of syntactic universals such as compositionality and Greenbergian word order universals~\citep{greenberg-1963} can be explained by this hypothesis~\citep{kirby-etal-2015,hahn-etal-2020}.

But we do not yet know whether this type of competition-based account can be extended to more abstract types of linguistic knowledge that go beyond mere sensitivity to structure such as compositionally and word order.
A representative type of such abstract knowledge comes from cases of what one might call \emph{structure dependence}, by which we broadly refer to operations that directly manipulate structural representations at some level of linguistic representation. 
Structure dependence has traditionally been taken to be a characteristic and central property of human language~\citep{chomsky-1957,chomsky-1965,everaert-etal-2015}; in fact, a key underlying theme throughout the whole history of mainstream generative grammar is extreme skepticism of the idea that such properties can be reduced to communicative principles.
The dogma in this line of thought has it that abstract syntactic properties of language are thought to be governed by domain-specific efficient computation necessary for deriving the structure of language~\citep{hauser-etal-2002,chomsky-2005,berwick-chomsky-2016}, while communication is viewed as \emph{not} essential to the core linguistic competence~\citep{chomsky-2002,hauser-etal-2002}.
It is thus crucial to investigate whether even such syntactic properties can be accounted for from the perspective of domain-general efficient communication.



In this paper, we directly address this issue by examining structure dependence.
Specifically, we investigate whether structure dependence realizes efficient communication by focusing on coordinate structures.
We design three types of languages: (i) one with a structure-dependent reduction operation, which has coordinate structures similar to those in natural language, (ii) one without any reduction operations, and (iii) one with a linear (rather than structure-dependent) reduction operation.
The latter two are conceptually possible but counterfactual languages.
We adopted~\citeposs{white-cotterell-2021} artificial probabilistic context-free grammars (PCFGs) to create the three languages.
Then we quantify the \emph{simplicity} and \emph{informativeness} of these languages and compare their communicative efficiency.
The results demonstrate that the languages with a structure-dependent reduction operation are significantly more communicatively efficient than their counterfactual counterparts.
This suggests that the structure-dependent properties in human language can be explained in terms of efficient communication.

\section{Background}
\subsection{Efficient communication hypothesis}\label{sec:effcomhypo}
In recent years, many researchers in cognitive science and computational psycholinguistics have increasingly focused on attributing cross-linguistic properties to domain-general cognitive functions.
The central thesis of this strand of research is that natural language is shaped to achieve efficient communication~\citep{zipf-1949,jaeger-tily-2011,christiansen-chater-2016,kemp-etal-2018,gibson-etal-2019,futrell-hahn-2022,fedorenko-etal-2024}.
Communicatively efficient structures are more likely to be learned because they may be used more frequently and are easier to process during learning.
This can drive changes in the language that further enhance communicative efficiency~\citep{jaeger-tily-2011,fedzechkina-etal-2012}.
Alternatively, the intergenerational transmission bottleneck in cultural evolution might lead to the selection of communicatively efficient languages~\citep{christiansen-kirby-2003,kirby-etal-2015}.
In either case, if functional pressures for efficient communication are at work, languages are expected to be optimized for efficient communication.

To test this hypothesis, one approach is to quantify and compare the communicative efficiency of real languages with logically possible but unattested counterfactual languages.
For example,~\citet{hahn-etal-2020} showed that real languages reach an optimal word order under the trade-off between simplicity and informativeness.
Simplicity refers to how simple the sentences of a language are as strings, while informativeness indicates how accurately the meaning can be reconstructed from the sentences of that language.
The communicative efficiency of a language is then defined as the weighted sum of simplicity and informativeness.
They created counterfactual languages for each of the 51 natural languages by changing word order patterns while maintaining the projectivity of dependency structures and calculated the communicative efficiency of all the languages.
They found that almost all of the real languages had significantly higher communicative efficiency than their counterparts.
This indicates that natural language is shaped by the pressure to enhance communicative efficiency.

\subsection{Structure dependence}
It has long been argued that natural language syntax exhibits structure dependence, the sensitivity to a hierarchical syntactic structure rather than a linear sequence of words~\citep{chomsky-1957,chomsky-1965,everaert-etal-2015}.
Grammatical operations are thus applied based on syntactic structures, not on linear strings.
For instance, yes-no questions in English are a well-known syntactic phenomenon that requires a structure-dependent grammatical rule.
In English yes-no questions, the auxiliary of the main clause moves to the front of the sentence.
If we were to formulate the rule for forming yes-no questions as \textit{move the leftmost auxiliary verb to the front}, which is not structure-dependent, we would incorrectly transform a sentence \textit{The man who is running is happy} into \textit{Is the man who running is happy?}
The correct transformation should move the second \textit{is} from the main clause, resulting in \textit{Is the man who is running happy?}

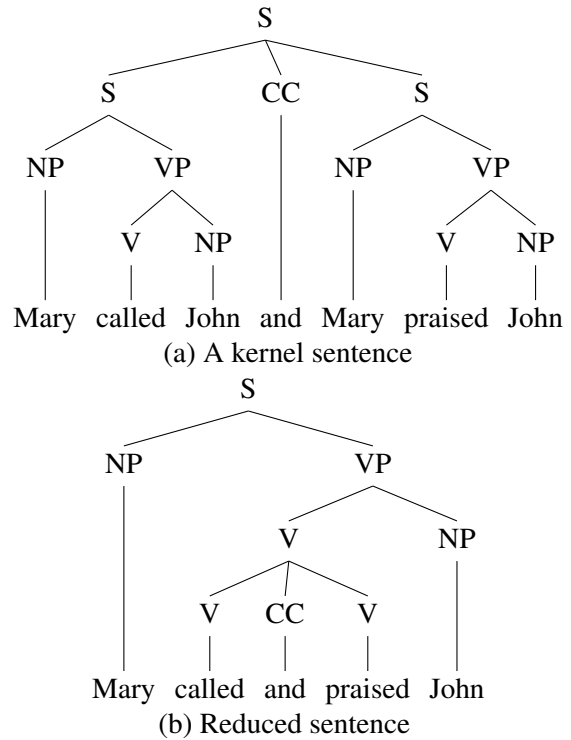
\begin{figure}[t]
    \centering
    \begin{subfigure}
        \centering
        \begin{forest}
            for tree={s sep=2mm, inner sep=0.2mm, l=0}
            [S [S [NP [Mary, tier=word]][VP [V [called, tier=word]][NP [John, tier=word]]]][CC [and, tier=word]][S [NP [Mary, tier=word]][VP [V [praised, tier=word]][NP [John, tier=word]]]]]
        \end{forest}
        \\
        (a) A kernel sentence
    \end{subfigure}
    \begin{subfigure}
        \centering
        \begin{forest}
            for tree={s sep=2mm, inner sep=0.2mm, l=0}
            [S [NP [Mary, tier=word]][VP [V [V [called, tier=word]][CC [and, tier=word]][V [praised, tier=word]]][NP [John, tier=word]]]]
        \end{forest}
        \\
        (b) Reduced sentence
    \end{subfigure}
    \caption{A coordinate structure (b) is derived by applying Conjunction Reduction, a structure-dependent reduction operation to a sentence-level coordinated kernel sentence (a).}
    \label{fig:coordinate_reduction}
\end{figure}

In the same way, it has traditionally been assumed that coordination, which is the focus of this study, also requires a structure-dependent grammatical operation~\citep{chomsky-1957,chomsky-1955,ross-1967}.
For example, the coordinated sentence \textit{John ran and swam} is derived from \textit{John ran and John swam}, and \textit{Mary called and praised John} is derived from \textit{Mary called John and Mary praised John}, through a structure-dependent grammatical operation known as \emph{Conjunction Reduction} (Figure~\ref{fig:coordinate_reduction}).
Conjunction Reduction is formulated as follows:
\begin{align*}
    & (Y+X_1+Z)+\text{CC}+(Y+X_2+Z) \\
    \to\ &Y+(X_1+\text{CC}+X_2)+Z, \\
    \tag{\citealp{chomsky-1957}, p.113, with slight modifications}
\end{align*}
where $X$ represents any syntactic category, $Y$ and $Z$ represent any syntactic category or string, CC represents any conjunction, and $+$ denotes concatenation.
Conjunction Reduction captures the fact that coordination is possible between identical syntactic categories for any syntactic category.\footnote{
    In the subsequent linguistic literature, Conjunction Reduction has been replaced by a more sophisticated approach known as \emph{Generalized Conjunction}~\citep{gazdar-1980,partee-rooth-1983}, which overcomes some important limitations of Conjunction Reduction~\citep[e.g.,][]{partee-1970}.
    However, the fundamental insight of structure dependence in the classical formulation of Conjunction Reduction is fully retained in Generalized Conjunction, since the latter can essentially be viewed as a reformulation of the former at the level of semantic representation using lambda calculus and higher-order functions.
    This study focuses on the operations applied to structures at any level.
    Therefore, the difference between Conjunction Reduction and Generalized Conjunction is orthogonal to the following discussion.
    }

We aim to investigate whether this structure-dependent reduction operation contributes to communicative efficiency in natural language.

\section{Experiment}
\subsection{Data}
\paragraph{Design of languages}
We designed the following three types of languages to investigate the impact on communicative efficiency when a language does not have a structure-dependent reduction operation at all:
\begin{enumerate}
    \item \texttt{No-reduction} language: A language with no reduction. Only sentence-level coordination is possible.
    \item \texttt{Structure-reduction} language: A language with structure-dependent reduction. Coordination is possible between identical syntactic categories.
    \item \texttt{Linear-reduction} language: A language with linear (rather than structure-dependent) reduction where repeated expressions in the same sentence are deleted in a coordinate structure.
\end{enumerate}

\paragraph{Data generation}
For each language, the sentences to be evaluated were created using a set of PCFGs defined by~\citet{white-cotterell-2021}.
The PCFGs are equipped with six switches to reverse the linear order of specific heads and dependents, which results in a total of $2^6=64$ word order patterns in the artificial languages.

\begin{table}[t]
    \centering
    \begin{tabular}{p{0.6cm}p{0.2cm}p{5.7cm}}
        \toprule
        \multicolumn{3}{l}{\textbf{CFG rules}} \\
        \midrule
        S & $\rightarrow$ & NP$_{\text{Subj}}$ VP \\
        VP & $\rightarrow$ & IVerb $|$ TVerb NP$_{\text{Obj}}$ $|$ Verb$_{\text{Comp}}$ S$_{\text{Comp}}$ \\
        S$_{\text{Comp}}$ & $\rightarrow$ & Comp S\\
        NP & $\rightarrow$ & Adj NP $|$ NP PP $|$ NP Rel VP \\
        NP$_{\text{Subj}}$ & $\rightarrow$ & Noun Case$_{\text{Subj}}$ $|$ Pronoun$_{\text{Subj}}$ \\
        NP$_{\text{Obj}}$ & $\rightarrow$ & Noun Case$_{\text{Obj}}$ $|$ Pronoun$_{\text{Obj}}$ \\
        PP & $\rightarrow$ & Prep NP \\ 
        $X$ & $\rightarrow$ & $X$ CC $X$, where $X=$\{NP, Adj, IVerb, TVerb\} \\
        \bottomrule
    \end{tabular}
    \caption{Overview of the grammatical rules equipped in \citeposs{white-cotterell-2021} PCFG.
            For simplicity, features such as tense and number are omitted.
            }
    \label{tab:cfg rules}
\end{table}

The PCFG includes the following basic syntactic categories: verb, noun, pronoun, adjective, conjunction, preposition, particle, sentential complementizer, and relativizer.
Some features such as tense (present and past), number (singular and plural), and grammatical relation (subject and object) are assigned to categories.
The grammatical rules are defined by the categories, as shown in Table~\ref{tab:cfg rules}.
The combination of categories and features results in a total of 44 syntactic categories and a lexicon consists of 1,254 words.
Although this grammar is much simpler than that of real natural languages, it is sufficiently sophisticated for our purpose of comparing structurally different languages.
Moreover, it allows us to simultaneously take into consideration typologically diverse word order patterns.

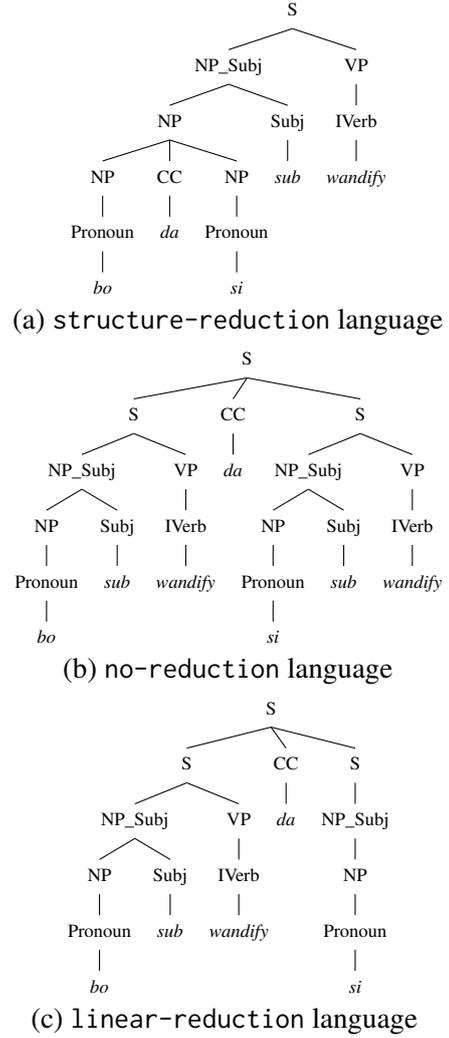
\begin{figure}[t]
    \centering
    \begin{subfigure}
        \centering
        {\scriptsize
        \begin{forest}
        [S [NP\_Subj [NP [NP [Pronoun [\textit{bo}]]] [CC [\textit{da}]] [NP [Pronoun [\textit{si}]]]] [Subj [\textit{sub}]]] [VP [IVerb [\textit{wandify}]]]]
        \end{forest}}\\
        (a) \texttt{structure-reduction} language
    \end{subfigure}
    \hfill
    \begin{subfigure}
        \centering
        {\scriptsize
        \begin{forest}
	      [S [S [NP\_Subj [NP [Pronoun [\textit{bo}]]] [Subj [\textit{sub}]]] [VP [IVerb [\textit{wandify}]]]] [CC [\textit{da}]] [S [NP\_Subj [NP [Pronoun [\textit{si}]]] [Subj [\textit{sub}]]] [VP [IVerb [\textit{wandify}]]]]]
        \end{forest}}\\
        (b) \texttt{no-reduction} language
    \end{subfigure}
    \hfill
    \begin{subfigure}
        \centering
        {\scriptsize
        \begin{forest}
        [S [S [NP\_Subj [NP [Pronoun [\textit{bo}]]] [Subj [\textit{sub}]]] [VP [IVerb [\textit{wandify}]]]] [CC [\textit{da}]] [S [NP\_Subj [NP [Pronoun [\textit{si}]]]]]]
        \end{forest}}\\
        (c) \texttt{linear-reduction} language
    \end{subfigure}
    \caption{Examples of the three languages expressing the same meaning. The word order is set with all six switches being strictly head-final as in Japanese. For simplicity, information on number and tense has been omitted from the syntactic categories in these figures.}
    \label{fig:artificial languages}
\end{figure}

We used this PCFG to create corpora of the artificial languages with 64 different word orders.
We then constructed the \texttt{no-reduction}, \texttt{structure-reduction}, and \texttt{linear-reduction} languages defined above for each of these word orders.
The \texttt{structure-reduction} language is the direct output of the PCFG.
We then expanded all of the coordinate structures in the \texttt{structure-reduction} language to a sentence level to create the \texttt{no-reduction} language.
Furthermore, we applied a linear reduction to the \texttt{no-reduction} language by deleting all repeated words in the same coordinate structure to create the \texttt{linear-reduction} language.
The examples of the tree structures of the three types of artificial languages are shown in Figure~\ref{fig:artificial languages}.

We measure the communicative efficiencies for these $64$ word orders $\times$ $3$ types  $= 192$ kinds of languages.

\subsection{Estimating communicative efficiency}
\paragraph{Definition of communicative efficiency}
For the simplicity/informativeness trade-off, following~\citet{hahn-etal-2020}, we evaluate simplicity as a property of the linear sequence, in terms of how easily the next word in an utterance can be predicted, i.e., \emph{predictability}, and informativeness in terms of how well the syntactic structure behind an utterance can be reconstructed, i.e., \emph{parsability}.

Predictability is specifically defined as the negative entropy, $-H(\mathcal{U})$, of all utterances $u$ in a language:
\begin{align}
-H(\mathcal{U}) = \sum_{u\in\mathcal{U}} p(u)\log p(u).
\end{align}
Here, we define the probability of utterance $u$ as the product of the probabilities of the words that constitute the utterance.
When the sample size is sufficiently large, entropy can be estimated as the mean word-by-word surprisal.
Surprisal is a metric that empirically predicts human behavioral~\citep[e.g.,][]{demberg-keller-2008,smith-levy-2013,shain-etal-2024} and neural~\citep[e.g.,][]{frank-etal-2015,lopopolo-etal-2017,brennan-hale-2019,shain-etal-2020} data.
The mean negative word-by-word surprisal represents the ease of incremental sentence processing on average under surprisal theory~\citep{hale-2001,levy-2008}.

Parsability is defined as the negative conditinal entropy, $-H(\mathcal{T}|\mathcal{U})$, of the underlying syntactic structure $t$ given an utterance $u$:\footnote{
\citet{hahn-etal-2020} cenceptually defined parsability as mutual information $I(\mathcal{U};\mathcal{T})=H(\mathcal{T})-H(\mathcal{T}|\mathcal{U})$ between an utterance and its syntactic structure.
However, they actually estimated the value of parsability as $-H(\mathcal{T}|\mathcal{U})$ on the assumption that $H(\mathcal{T})$ is constant.}
\begin{align}
-H(\mathcal{T}|\mathcal{U}) = \sum_{t\in\mathcal{T},u\in\mathcal{U}}p(t,u)\log p(t|u).
\end{align}
Since semantic calculation in compositional semantics crucially depends on the building of syntactic structures~\citep{montague-1970,heim-kratzer-1998}, we employ a metric of informativeness that captures how unambiguously the underlying syntactic structure can be reconstructed---both temporally and globally---as an indicator of how accurately the intended meanings of utterances can be recovered.

Then, following~\citet{ferrer-i-cancho-sole-2003} and~\citet{hahn-etal-2020}, we defined a communicative efficiency function as the weighted sum of predictability and parsability:
\begin{align}
\Omega(\lambda) &\coloneqq \lambda \text{predictability} + (1-\lambda) \text{parsability} \\
&= -\lambda H(\mathcal{U}) - (1-\lambda)H(\mathcal{T}|\mathcal{U}),\label{eq:communicative efficiency}
\end{align}
where $\lambda$ is a trade-off parameter ranging from $0$ to $1$, which represents the contribution of each term.
The objective function that captures the trade-off between the cost of linguistic expressions and the likelihood of meaning given the expression has been used in previous studies~\citep[e.g.,][]{ferrer-i-cancho-sole-2003,frank-goodman-2012,kemp-regier-2012,regier-etal-2015,hahn-etal-2020}.

\paragraph{Recurrent Neural Network Grammars}
To obtain the values of predictability and parsability, we adopted Recurrent Neural Network Grammars (RNNGs;~\citealp{dyer-etal-2016-recurrent}).
RNNGs are a generative model of sentences that explicitly models hierarchical structures by processing the action sequence of shift-reduce parsing.
RNNGs can be used for both language modeling and parsing with the same model parameters, which is suitable for our purpose here.

In this study, we used the left-corner stack-only RNNGs~\citep{kuncoro-etal-2018} implemented with PyTorch\footnote{\url{https://github.com/pytorch/pytorch/releases/tag/v1.12.1}} by~\citet{noji-oseki-2021}.
We used a two-layer LSTM, where both the hidden layer and the input layer have 256 dimensions.\footnote{
Other hyperparameters are as follows:
random seeds are $\{3435, 3436, 3437\}$, optimizer is Adam ~\citep{kingma-ba-2015}, learning rate is $0.001$, dropout is $0.3$, and batch size is $128$.
The code of RNNGs we employed~\citep{noji-oseki-2021} is available at~\url{https://github.com/aistairc/ rnng-pytorch}.
}

Left-corner parsing is considered reasonable for human incremental sentence processing from the perspective of memory capacity~\citep{abney-johnson-1991,resnik-1992} and is often assumed as a model of human sentence processing~\citep[e.g.,][]{lewis-vasishth-2005,van-schijndel-etal-2013}.
Additionally, it has already been pointed out that a simple bottom-up strategy without a predictive process cannot explain the human incremental processing of coordinate structures in English at least when the parser conducts a serial parsing~\citep{sturt-lombardo-2005,stanojevic-etal-2023}.
This motivates our choice of a left-corner as a parsing strategy.

We performed a beam search with a beam size of 100 for inference.
We also used word-synchronous beam search~\citep{stern-etal-2017} with a size of 10.

\paragraph{Estimation of communicative efficiency}
Predictability for the languages can be obtained by
\begin{align}
-H(\mathcal{U}) = \sum_{u\in\mathcal{U}} p(u)\log p_\phi(u),    
\end{align}
and following~\citet{hahn-etal-2020}, we define the log-likelihood of each utterance $u$ as 
\begin{align}
\log p_\phi(u) \coloneqq \sum^N_{i=1}\log p_\phi (w_i|w_{<i}),    
\end{align}
where $w_i$ represents the $i$-th word composing the utterance and $\phi$ represents the parameters of the RNNGs.
We can approximate the negative entropy of $u$ with its Monte Carlo estimate on test data:
\begin{align}
    -H(\mathcal{U}) \approx \frac{1}{|\text{Test Data}|}\sum_{u\in\text{Test Data}} \log p_\phi(u).
\end{align}
We calculated the values of predictability according to the formula above.

Parsability can be calculated by
\begin{align}
-H(\mathcal{T}|\mathcal{U}) = \sum_{t\in\mathcal{T},u\in\mathcal{U}}p(t,u)\log p_\phi (t|u),
\end{align}
and in the same way, the conditional entropy can be approximated with its Monte Carlo estimate on test data:
\begin{align}
    -H(\mathcal{T}|\mathcal{U}) \approx \frac{1}{|\text{Test Data}|}\sum_{t,u\in\text{Test Data}} \log p_\phi (t|u).
\end{align}
Here, we define the log-likelihood of the conditional probability of the tree structure $t$ given each utterance $u$ as
\begin{align}
\log p_\phi (t|u) \coloneqq \sum^N_{i=1}\log p_\phi(t_{\text{best}}|w_{\leq i}).   
\end{align}
Again, $w_i$ and $\phi$ represent the $i$-th word of the utterance and the parameters of RNNGs, respectively.
$t_{\text{best}}$ refers to the most likely constituency parse in the word-synchronous beam at each word.

For 192 types of artificial languages, we generated 20,000 sentences for each and divided them into an 8-1-1 train-dev-test split for training and evaluation.
For all languages, we trained RNNGs on word-by-word using Adam~\citep{kingma-ba-2015} for 10 epochs each with multiple random seeds.
Then, we calculated the values of predictability and parsability, normalized by the number of words, to ensure valid comparisons across languages with inherently different sentence lengths.

\section{Results}
\begin{figure*}[t]
    \centering
    \includegraphics[scale=0.6]{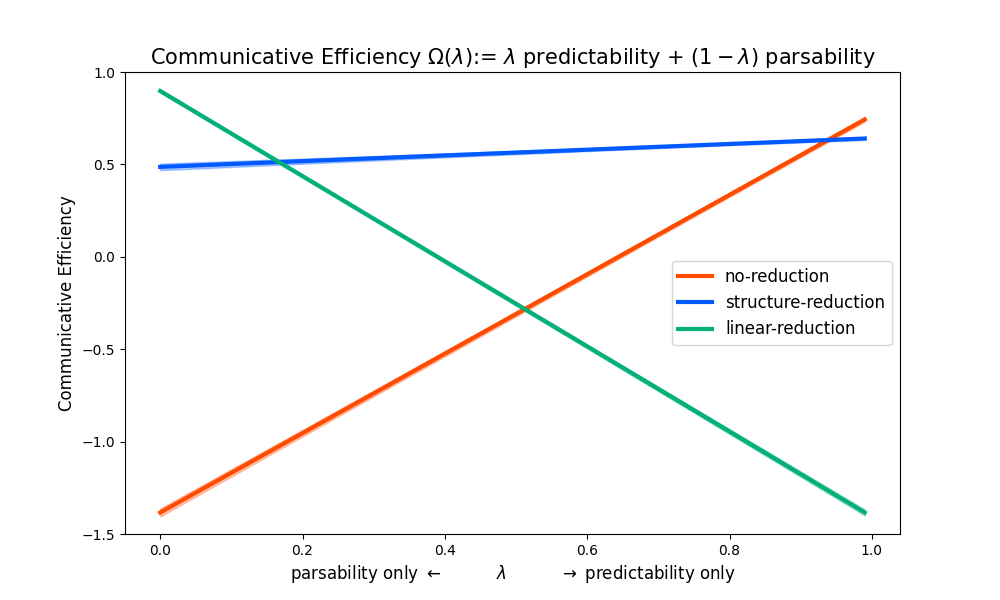}
    \caption{Distribution of \textbf{communicative efficiency} for the three types of languages with 95\% CI.
    The \textit{x}-axis and \textit{y}-axis represent the trade-off parameter $\lambda$ and communicative efficiency, respectively.
    Both predictability and parsability are \textit{z}-transformed for an interpretation of $\lambda$.
    The \texttt{structure-reduction} languages are the most communicatively efficient under the parameter $\lambda\in[0.18, 0.93]$ for 95\% CI.}
    \label{fig:communicative_efficiency}
\end{figure*}

A distribution of the values of communicative efficiency for the three types of languages, as calculated by RNNGs, is shown in Figure~\ref{fig:communicative_efficiency}.
For an interpretation of the trade-off parameter $\lambda$, the predictability and parsability values of all languages are \textit{z}-transformed (i.e., centered and divided by the standard deviation) before being substituted into Eq~\ref{eq:communicative efficiency}.
The lines in the figure show the transitions of the value of communicative efficiency for $\lambda$ with a 95\% confidence interval (CI).
By finding the coordinates where the lines intersect, we can observe the behavior of communicative efficiency for each language depending on the value of $\lambda$.
The lower bound of the 95\% CI for \texttt{structure-reduction} intersects with the upper bound of the 95\% CI for \texttt{linear-reduction} at $\lambda=0.18$.
In the same way, the upper bound of the 95\% CI for \texttt{structure-reduction} intersects with the lower bound of the 95\% CI for \texttt{linear-reduction} at $\lambda=0.93$.
This indicates that \texttt{structure-reduction} languages are the most communicatively efficient, at least within the range of $\lambda$ values between 0.18 and 0.93.

\begin{figure}[t]
  \centering
  \includegraphics[scale=0.6]{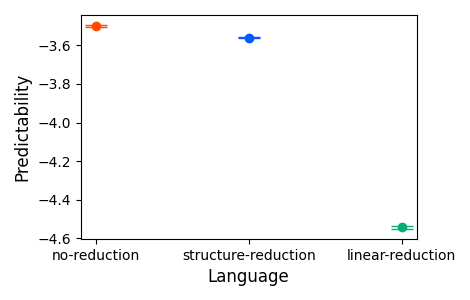}
  \caption{Distribution of \textbf{predictability} for the three types of languages.
  Error bars indicate 95\% CI.}
  \label{fig:predictability}
\end{figure}
\begin{figure}
  \centering
  \includegraphics[scale=0.6]{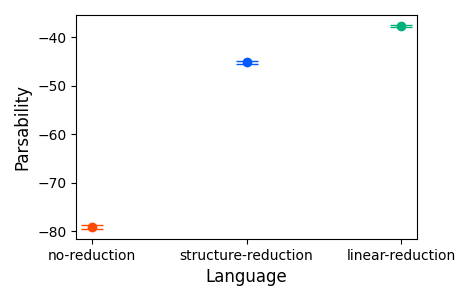}
  \caption{Distribution of \textbf{parsability} for the three types of languages.
  Error bars indicate 95\% CI.}
  \label{fig:parsability}
\end{figure}

\begin{figure}[t]
    \centering
    \subfigure[Relationship between \textbf{predictability} and word position.]
    {
        \includegraphics[scale=0.6]{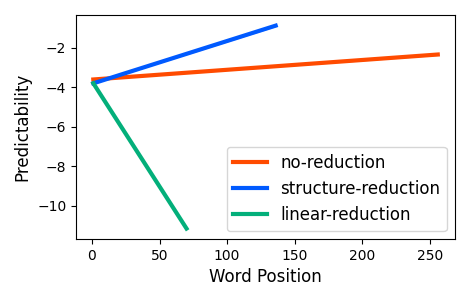}
        \label{fig:regr_pred}
    }
    \\
    \subfigure[Relationship between \textbf{parsability} and word position.]
    {
        \includegraphics[scale=0.6]{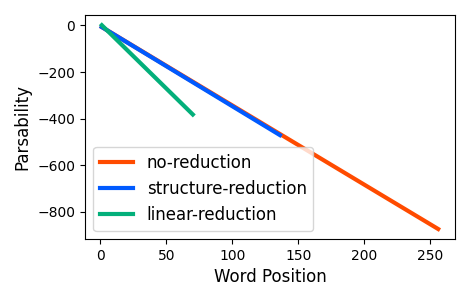}
        \label{fig:regr_parse}
    }
    \caption{Relationship between predictability/parsability and word position for the three types of languages.
    Predictability and parsability here refer to the negative surprisal and the negative log-likelihood of the best parse for each word, respectively.
    The lines represent the fit of a least squares regression model for these data.}
    \label{fig:regr}
\end{figure}


Additionally, when we examine only predictability or parsability, their distributions are shown in Figure~\ref{fig:predictability} and Figure~\ref{fig:parsability}, respectively.
The mean values for predictability were ordered as \texttt{no-reduction} $>$ \texttt{structure-reduction} $>$ \texttt{linear-reduction}, while for parsability, the order was \texttt{linear-reduction} $>$ \texttt{structure-reduction} $>$ \texttt{no-reduction}, in which all of the pairs have a statistically significant difference ($p < 0.05/3$ by paired \textit{t}-test with Bonferroni correction).
This indicates that when only predictability or parsability is individually considered, the \texttt{structure-reduction} language may not always be the best option.
However, to satisfy both criteria simultaneously, i.e., to consider the weighted sum of predictability and parsability under the parameter $\lambda\in[0.18, 0.93]$, the \texttt{structure-reduction} language achieves the highest score of communicative efficiency as shown in Figure~\ref{fig:communicative_efficiency}.

To further interpret the results of predictability and parsability, we plotted least squares regression lines between word position in the sentence and each word-by-word value (Figure~\ref{fig:regr}).
For predictability, only the \texttt{linear-reduction} languages significantly decline towards the latter part of the sentence.
As for parsability, although the \texttt{linear-reduction} languages experience a faster decrease, the other two languages, which have longer expression lengths, achieve a lower overall score.

\section{Discussion}
We demonstrated that the \texttt{structure-reduction} languages, which have the same structure-dependent reduction operation as natural language, had significantly higher communicative efficiency compared to the conceptually possible but counterfactual \texttt{no-reduction} and \texttt{linear-reduction} languages when we calculated the scores by RNNGs with a trade-off parameter $\lambda\in[0.18, 0.93]$.
This suggests that the structure-dependent reduction operation prevalent in the natural language syntax may exist due to functional pressures to support efficient communication along lines discussed in Section~\ref{sec:effcomhypo}.
It should be noted that, as Figure~\ref{fig:communicative_efficiency} shows, when $\lambda$ is extremely small or large, the \texttt{no-reduction} or \texttt{linear-reduction} languages achieve the highest efficiency scores.
However, $\lambda$ represents the relative contribution of the two terms to the overall efficiency score.
It is difficult to assume a reasonable scenario where only one term is emphasized.
While we do not aim to estimate a specific value of $\lambda$, empirically,~\citet{ferrer-i-cancho-sole-2003} found in their simulation experiment that Zipf's law emerged when $\lambda\approx 0.41$.
In addition,~\citet{hahn-etal-2020} demonstrated that grammars optimized at $\lambda\approx 0.47$ captured all 8 of the Greenberg correlations they investigated, whereas optimizing solely for predictability or parsability did not account for all of them.\footnote{
\citeauthor{hahn-etal-2020} defined a communicative efficiency function as $\Omega(\lambda)\coloneqq\lambda\text{predictability}+\text{parsability}$ and set $\lambda=0.9$.
In other words, while we assigned weights of $\lambda$ and $1-\lambda$ to predictability and parsability, respectively, they assigned weights of $0.9$ and $1$.
Solving the equation $\lambda/(1-\lambda) = 0.9/1$ gives $\lambda\approx 0.47$.
}

When we consider only one of the terms constituting communicative efficiency, that is, predictability (simplicity) or parsability (informativeness), one of the two counterfactual languages achieves the highest score.
The \texttt{no-reduction} language is the easiest in terms of prediction, i.e., the estimation of the next word, among the three types.
This is because the language has no reduction, and all sentences are fully represented, so the number of local patterns for the next word is limited, which makes prediction easier.
For example, in a \texttt{no-reduction} language where the part of speech at the beginning of a sentence is always $X$ in its word order pattern, the part of speech following a conjunction must be $X$, while \texttt{structure-reduction} and \texttt{linear-reduction} languages have more variations, which lead to higher entropy of strings.
However, the \texttt{no-reduction} language is not well-suited for parsing.
As sentence length increases, the number of potential parses grows exponentially~\citep{church-patil-1982}.
Since this language lacks reduction and results in longer overall expressions, the possible parses at each word position rapidly increase, as shown in Figure~\ref{fig:regr_parse}.
Consequently, it is not an optimal design from the perspective of estimating the underlying structure of an utterance.
In contrast, the \texttt{linear-reduction} language has shorter overall expressions, resulting in fewer possible parses at each word position.\footnote{
The parsability metric used here may overestimate the informativeness of \texttt{linear-reduction} languages.
This metric does not account for whether the estimated tree structure is correct, nor does it fully capture the inherent ambiguity in the language.
A more accurate approach could involve quantifying informativeness using mutual information~\citep[e.g.,][]{ferrer-i-cancho-2005,futrell-2017,zaslavsky-etal-2018,hahn-etal-2020}, though this would not change our conclusion.
}
As a result, it is superior in terms of estimating tree structures.
However, the reduction is too radical, making it challenging to maintain predictability.
As shown in Figure~\ref{fig:regr_pred}, this issue is evident in the latter parts of sentences in this language, where predictions become increasingly difficult due to the need to consider the possibility that previously mentioned words might have been reduced.
In short, a reduction operation is necessary to enhance parsability, but it should be applied \emph{restrictively} so as not to sacrifice next-word predictability.
Balancing the trade-off between the two, a structure-dependent reduction is the most preferred design for maximizing communicative efficiency.

It should be noted that even in natural language, there are instances of linear reduction operations such as \emph{stripping}~\citep{hankamer-sag-1976}, for instance shown below, or sentences that retain sentence-level coordination without reduction for pragmatic purposes such as emphasis.
\ex.[(11)] Mary took a walk in the park, and Bill too.

However, these phenomena occur as alternative choices in a language that has a structure-dependent reduction operation.
Our claim is that a language lacking a structure-dependent reduction operation entirely is not preferable from the perspective of efficient communication.

The results of this study have interesting implications for theoretical linguistics research.
Structure dependence, a syntactic universal property we addressed here, has traditionally been argued to be one of the characteristic features of human language~\citep{chomsky-1957,chomsky-1965,everaert-etal-2015}.
A prominent view in the mainstream generative grammar argues that natural language involves domain-specific predispositions and that syntactic properties of language---including structure dependence---are best explained from the perspective of `efficient \emph{computation}' reflecting such predispositions genetically hard-wired in the human brain~\citep{hauser-etal-2002,chomsky-2005,everaert-etal-2015,berwick-chomsky-2016}.
Under this view, communication is taken to be a kind of epiphenomenon, not essential to the core linguistic competence~\citep{chomsky-2002,hauser-etal-2002}.
However, our results suggest that at least some structure-dependent properties present in natural language (such as coordination) can be explained from the perspective of efficient \emph{communication}.
This does not immediately refute the dominant research program attempting to explain linguistic properties from a `computational' perspective, but it does indicate that abstract properties in syntax may not necessarily need to be explained solely from that perspective.
This aligns with the existing body of research that attempts to explain various aspects of natural language from the perspective of efficient communication~\citep{gibson-etal-2019,fedorenko-etal-2024}.

\section{Conclusion}
In this paper, we investigated whether structure dependence, one of the syntactic universals, reflects the optimization for efficient communication.
To address this issue, we focused on coordinate structures and designed three types of artificial languages: (i) one with a structure-dependent reduction operation, (ii) one without any reduction operations, and (iii) one with a linear (rather than structure-dependent) reduction operation.
We quantified the communicative efficiency of these languages and compared them.
The results demonstrated that the languages with a structure-dependent reduction operation were significantly more communicatively efficient than their counterfactual counterparts.
This suggests that the structure-dependent properties of natural language can be explained from the functional perspective of efficient communication.\footnote{Code for reproducing our experiments is available at \url{https://github.com/kohei-kaji/coordination}.}

\section*{Limitations}

There is room for improvement in the objective function for communicative efficiency.
Although we used the mean word-by-word surprisal, conditioned on all preceding words, as a measure of predictability, human language processing is subject to short-term memory constraints~\citep{gibson-1998,lewis-vasishth-2005,isono-2024}.
Thus, it is preferable to model predictability in a way that incorporates \emph{lossy memory representation}~\citep{futrell-etal-2020-lossy,hahn-etal-2021,hahn-etal-2022-resource}.
Moreover, the psychological plausibility of the parsability metric should be critically evaluated, both conceptually and empirically.
Since parsability relies on an intermediate representation---syntactic structures---it does not fully capture the direct relationship between linguistic expressions and their meanings, suggesting that there is room for further conceptual refinement.

To the best of our knowledge, this study is the first to investigate communicative efficiency with respect to structure dependence.
We focused on coordinate structures with the artificial language paradigm as a starting point.
Of course, a deeper understanding of structure dependence in language will require using natural language data and extending the analysis to phenomena such as agreement and movement that are argued to be relevant to structure dependence.
In future work, we plan to test the relationship between structure dependence and communicative efficiency by applying the methodology proposed here to a broader range of syntactic constructions, using treebanks like Universal Dependencies~\citep{nivre-etal-2020}.


\section*{Ethical considerations}
We used all tools and datasets following their respective terms and licenses.
We employed ChatGPT and Grammarly for writing assistance and utilized ChatGPT for writing experimental code.
We used these tools in compliance with the ACL 2023 Policy on AI Writing Assistance.

\section*{Acknowledgments}
We are grateful to the anonymous reviewers for their insightful comments and suggestions.
We also thank Ryo Yoshida for providing the code to calculate the log-likelihood of each parse in RNNGs.
We sincerely appreciate Shinnosuke Isono, Douglas Roland, Yushi Sugimoto, Asa Tomita, and members of the computational linguistics community at The University of Tokyo for their valuable feedback.
This work was supported by JSPS KAKENHI Grant Numbers 21K00541 and 24H00087, JST PRESTO Grant Number JPMJPR21C2, and the NINJAL collaborative research project `Toward a Computationally-Informed Theoretical Linguistics'.

\bibliography{custom}

\begin{thebibliography}{79}
\providecommand{\natexlab}[1]{#1}

\bibitem[{Abney and Johnson(1991)}]{abney-johnson-1991}
Steven~P. Abney and Mark Johnson. 1991.
\newblock \href {https://doi.org/https://psycnet.apa.org/doi/10.1007/BF01067217} {Memory requirements and local ambiguities of parsing strategies}.
\newblock \emph{Journal of Psycholinguistic Research}, 20(3):233--250.

\bibitem[{Berwick and Chomsky(2016)}]{berwick-chomsky-2016}
Robert~C. Berwick and Noam Chomsky. 2016.
\newblock \href {https://mitpress.mit.edu/9780262533492/why-only-us/} {\emph{Why Only Us: Language and Evolution}}.
\newblock The MIT Press, Cambridge, MA.

\bibitem[{Brennan and Hale(2019)}]{brennan-hale-2019}
Jonathan~R. Brennan and John~T. Hale. 2019.
\newblock \href {https://doi.org/10.1371/journal.pone.0207741} {Hierarchical structure guides rapid linguistic predictions during naturalistic listening}.
\newblock \emph{PLoS ONE}, 14(1):e0207741.

\bibitem[{Chen et~al.(2023)Chen, Futrell, and Mahowald}]{chen-etal-2023}
Sihan Chen, Richard Futrell, and Kyle Mahowald. 2023.
\newblock \href {https://doi.org/10.1016/j.cognition.2023.105505} {An information-theoretic approach to the typology of spatial demonstratives}.
\newblock \emph{Cognition}, 240:105505.

\bibitem[{Chomsky(1957)}]{chomsky-1957}
Noam Chomsky. 1957.
\newblock \href {https://doi.org/10.1515/9783112316009} {\emph{Syntactic Structures}}.
\newblock Mouton.

\bibitem[{Chomsky(1965)}]{chomsky-1965}
Noam Chomsky. 1965.
\newblock \href {https://www.jstor.org/stable/j.ctt17kk81z} {\emph{Aspects of the Theory of Syntax}}.
\newblock MIT Press.

\bibitem[{Chomsky({1975 (=1955)})}]{chomsky-1955}
Noam Chomsky. {1975 (=1955)}.
\newblock \href {https://link.springer.com/book/9780306307607} {\emph{The Logical Structure of Linguistic Theory}}.
\newblock Springer New York, NY.

\bibitem[{Chomsky(2002)}]{chomsky-2002}
Noam Chomsky. 2002.
\newblock \href {https://doi.org/10.1017/CBO9780511613876.005} {An interview on minimalism}.
\newblock In Adriana Belletti and Luigi Rizzi, editors, \emph{On Nature and Language}, pages 92--161. Cambridge University Press.

\bibitem[{Chomsky(2005)}]{chomsky-2005}
Noam Chomsky. 2005.
\newblock \href {https://doi.org/10.1162/0024389052993655} {{Three Factors in Language Design}}.
\newblock \emph{Linguistic Inquiry}, 36(1):1--22.

\bibitem[{Christiansen and Chater(2016)}]{christiansen-chater-2016}
Morten~H. Christiansen and Nick Chater. 2016.
\newblock \href {https://doi.org/10.1017/S0140525X1500031X} {{The Now-or-Never bottleneck: A fundamental constraint on language}}.
\newblock \emph{Behavioral and Brain Sciences}, 39:e62.

\bibitem[{Christiansen and Kirby(2003)}]{christiansen-kirby-2003}
Morten~H. Christiansen and Simon Kirby. 2003.
\newblock \href {https://doi.org/10.1016/S1364-6613(03)00136-0} {Language evolution: consensus and controversies}.
\newblock \emph{Trends in Cognitive Sciences}, 7(7):300--307.

\bibitem[{Church and Patil(1982)}]{church-patil-1982}
Kenneth Church and Ramesh Patil. 1982.
\newblock \href {https://aclanthology.org/J82-3004} {Coping with syntactic ambiguity or how to put the block in the box on the table}.
\newblock \emph{American Journal of Computational Linguistics}, 8(3-4):139--149.

\bibitem[{Clark et~al.(2023)Clark, Meister, Pimentel, Hahn, Cotterell, Futrell, and Levy}]{clark-etal-2023}
Thomas~Hikaru Clark, Clara Meister, Tiago Pimentel, Michael Hahn, Ryan Cotterell, Richard Futrell, and Roger Levy. 2023.
\newblock \href {https://doi.org/10.1162/tacl_a_00589} {A cross-linguistic pressure for {U}niform {I}nformation {D}ensity in word order}.
\newblock \emph{Transactions of the Association for Computational Linguistics}, 11:1048--1065.

\bibitem[{Demberg and Keller(2008)}]{demberg-keller-2008}
Vera Demberg and Frank Keller. 2008.
\newblock \href {https://doi.org/10.1016/j.cognition.2008.07.008} {Data from eye-tracking corpora as evidence for theories of syntactic processing complexity}.
\newblock \emph{Cognition}, 109(2):193--210.

\bibitem[{Deni{\'c} et~al.(2022)Deni{\'c}, Steinert-Threlkeld, and Szymanik}]{denic-etal-2022}
Milica Deni{\'c}, Shane Steinert-Threlkeld, and Jakub Szymanik. 2022.
\newblock \href {https://doi.org/10.1111/cogs.13142} {Indefinite pronouns optimize the simplicity/informativeness trade-off}.
\newblock \emph{Cognitive Science}, 46(5):e13142.

\bibitem[{Deni{\'c} and Szymanik(2024)}]{denic-szymanik-2024}
Milica Deni{\'c} and Jakub Szymanik. 2024.
\newblock \href {https://doi.org/10.1111/cogs.13424} {Recursive numeral systems optimize the trade-off between lexicon size and average morphosyntactic complexity}.
\newblock \emph{Cognitive Science}, 48(3):e13424.

\bibitem[{Dyer et~al.(2016)Dyer, Kuncoro, Ballesteros, and Smith}]{dyer-etal-2016-recurrent}
Chris Dyer, Adhiguna Kuncoro, Miguel Ballesteros, and Noah~A. Smith. 2016.
\newblock \href {https://doi.org/10.18653/v1/N16-1024} {Recurrent neural network grammars}.
\newblock In \emph{Proceedings of the 2016 Conference of the North {A}merican Chapter of the Association for Computational Linguistics: Human Language Technologies}, pages 199--209, San Diego, California. Association for Computational Linguistics.

\bibitem[{Everaert et~al.(2015)Everaert, Huybregts, Chomsky, Berwick, and Bolhuis}]{everaert-etal-2015}
Martin~B.H. Everaert, Marinus~A.C. Huybregts, Noam Chomsky, Robert~C. Berwick, and Johan~J. Bolhuis. 2015.
\newblock \href {https://doi.org/10.1016/j.tics.2015.09.008} {Structures, not strings: Linguistics as part of the cognitive sciences}.
\newblock \emph{Trends in Cognitive Sciences}, 19:729--743.

\bibitem[{Fedorenko et~al.(2024)Fedorenko, Piantadosi, and Gibson}]{fedorenko-etal-2024}
Evelina Fedorenko, Steven~T. Piantadosi, and Edward A.~F. Gibson. 2024.
\newblock \href {https://doi.org/10.1038/s41586-024-07522-w} {Language is primarily a tool for communication rather than thought}.
\newblock \emph{Nature}, 630:575--586.

\bibitem[{Fedzechkina et~al.(2012)Fedzechkina, Jaeger, and Newport}]{fedzechkina-etal-2012}
Maryia Fedzechkina, T.~Florian Jaeger, and Elissa~L. Newport. 2012.
\newblock \href {https://doi.org/10.1073/pnas.1215776109} {Language learners restructure their input to facilitate efficient communication}.
\newblock \emph{Proceedings of the National Academy of Sciences}, 109(44):17897--17902.

\bibitem[{{Ferrer i Cancho}(2005)}]{ferrer-i-cancho-2005}
Ramon {Ferrer i Cancho}. 2005.
\newblock \href {https://doi.org/10.1140/epjb/e2005-00340-y} {Zipf's law from a communicative phase transition}.
\newblock \emph{The European Physical Journal B - Condensed Matter and Complex Systems}, 47:449--457.

\bibitem[{{Ferrer i Cancho} and Sol{\'e}(2003)}]{ferrer-i-cancho-sole-2003}
Ramon {Ferrer i Cancho} and Ricard~V. Sol{\'e}. 2003.
\newblock \href {https://doi.org/10.1073/pnas.0335980100} {Least effort and the origins of scaling in human language}.
\newblock \emph{Proceedings of the National Academy of Sciences}, 100(3):788--791.

\bibitem[{Frank and Goodman(2012)}]{frank-goodman-2012}
Michael~C. Frank and Noah~D. Goodman. 2012.
\newblock \href {https://doi.org/10.1126/science.1218633} {Predicting pragmatic reasoning in language games}.
\newblock \emph{Science}, 336(6084):998--998.

\bibitem[{Frank et~al.(2015)Frank, Otten, Galli, and Vigliocco}]{frank-etal-2015}
Stefan~L. Frank, Leun~J. Otten, Giulia Galli, and Gabriella Vigliocco. 2015.
\newblock \href {https://doi.org/10.1016/j.bandl.2014.10.006} {The {ERP} response to the amount of information conveyed by words in sentences}.
\newblock \emph{Brain and Language}, 140:1--11.

\bibitem[{Futrell(2017)}]{futrell-2017}
Richard Futrell. 2017.
\newblock \emph{Memory and Locality in Natural Language}.
\newblock Ph.D. thesis, MIT.

\bibitem[{Futrell et~al.(2020{\natexlab{a}})Futrell, Gibson, and Levy}]{futrell-etal-2020-lossy}
Richard Futrell, Edward Gibson, and Roger~P Levy. 2020{\natexlab{a}}.
\newblock \href {https://onlinelibrary.wiley.com/doi/full/10.1111/cogs.12814} {Lossy-context surprisal: An information-theoretic model of memory effects in sentence processing}.
\newblock \emph{Cognitive Science}, 44(3):e12814.

\bibitem[{Futrell and Hahn(2022)}]{futrell-hahn-2022}
Richard Futrell and Michael Hahn. 2022.
\newblock \href {https://doi.org/10.3389/fcomm.2022.657725} {Information theory as a bridge between language function and language form}.
\newblock \emph{Frontiers in Communication}, 7.

\bibitem[{Futrell et~al.(2020{\natexlab{b}})Futrell, Levy, and Gibson}]{futrell-etal-2020-dependency}
Richard Futrell, Roger~P. Levy, and Edward Gibson. 2020{\natexlab{b}}.
\newblock \href {https://muse.jhu.edu/pub/24/article/757632} {Dependency locality as an explanatory principle for word order}.
\newblock \emph{Language}, 96(2):371--412.

\bibitem[{Gazdar(1980)}]{gazdar-1980}
Gerald Gazdar. 1980.
\newblock \href {https://doi.org/10.1007/BF00401693} {A cross-categorial semantics for coordination}.
\newblock \emph{Linguistics and Philosophy}, 3(3):407--409.

\bibitem[{Gibson(1998)}]{gibson-1998}
Edward Gibson. 1998.
\newblock \href {https://doi.org/10.1016/S0010-0277(98)00034-1} {Linguistic complexity: locality of syntactic dependencies}.
\newblock \emph{Cognition}, 68(1):1--76.

\bibitem[{Gibson et~al.(2019)Gibson, Futrell, Piantadosi, Dautriche, Mahowald, Bergen, and Levy}]{gibson-etal-2019}
Edward Gibson, Richard Futrell, Steven~T. Piantadosi, Isabelle Dautriche, Kyle Mahowald, Leon Bergen, and Roger Levy. 2019.
\newblock \href {https://doi.org/10.1016/j.tics.2019.02.003} {How efficiency shapes human language}.
\newblock \emph{Trends in Cognitive Sciences}, 23(5):389--407.

\bibitem[{Gildea and Jaeger(2015)}]{gildea-jaeger-2015}
Daniel Gildea and T.~Florian Jaeger. 2015.
\newblock \href {https://arxiv.org/abs/1510.02823} {Human languages order information efficiently}.
\newblock \emph{Preprint}, arXiv:1510.02823.

\bibitem[{Greenberg(1963)}]{greenberg-1963}
Joseph~H. Greenberg. 1963.
\newblock \emph{Universals of language}.
\newblock MIT press.

\bibitem[{Hahn et~al.(2021)Hahn, Degen, and Futrell}]{hahn-etal-2021}
Michael Hahn, Judith Degen, and Richard Futrell. 2021.
\newblock \href {https://doi.org/10.1037/rev0000269} {Modeling word and morpheme order in natural language as an efficient tradeoff of memory and surprisal}.
\newblock \emph{Psychological Review}, 128:726--756.

\bibitem[{Hahn et~al.(2022)Hahn, Futrell, Levy, and Gibson}]{hahn-etal-2022-resource}
Michael Hahn, Richard Futrell, Roger Levy, and Edward Gibson. 2022.
\newblock \href {https://doi.org/10.1073/pnas.2122602119} {A resource-rational model of human processing of recursive linguistic structure}.
\newblock \emph{Proceedings of the National Academy of Sciences}, 119(43):e2122602119.

\bibitem[{Hahn et~al.(2020)Hahn, Jurafsky, and Futrell}]{hahn-etal-2020}
Michael Hahn, Dan Jurafsky, and Richard Futrell. 2020.
\newblock \href {https://doi.org/10.1073/pnas.1910923117} {Universals of word order reflect optimization of grammars for efficient communication}.
\newblock \emph{Proceedings of the National Academy of Sciences}, 117(5):2347--2353.

\bibitem[{Hale(2001)}]{hale-2001}
John Hale. 2001.
\newblock \href {https://aclanthology.org/N01-1021} {A probabilistic {E}arley parser as a psycholinguistic model}.
\newblock In \emph{Second Meeting of the North {A}merican Chapter of the Association for Computational Linguistics}.

\bibitem[{Hankamer and Sag(1976)}]{hankamer-sag-1976}
Jorge Hankamer and Ivan~A. Sag. 1976.
\newblock Deep and surface anaphora.
\newblock \emph{Linguistic Inquiry}, 7:391--428.

\bibitem[{Haspelmath(2008)}]{haspelmath-2008}
Martin Haspelmath. 2008.
\newblock \href {https://doi.org/10.1075/la.132.04has} {Parametric versus functional explanations of syntactic universals}.
\newblock \emph{The Limits of Syntactic Variation}, 132:75--107.

\bibitem[{Hauser et~al.(2002)Hauser, Chomsky, and Fitch}]{hauser-etal-2002}
{Marc D.} Hauser, Noam Chomsky, and {W. Tecumseh} Fitch. 2002.
\newblock \href {https://doi.org/10.1126/science.298.5598.1569} {The faculty of language: What is it, who has it, and how did it evolve?}
\newblock \emph{Science}, 298(5598):1569--1579.

\bibitem[{Hawkins(1994)}]{hawkins-1994}
John~A. Hawkins. 1994.
\newblock \href {https://doi.org/10.1017/CBO9780511554285} {\emph{A Performance Theory of Order and Constituency}}.
\newblock Cambridge University Press.

\bibitem[{Hawkins(2004)}]{hawkins-2004}
John~A. Hawkins. 2004.
\newblock \href {https://doi.org/10.1093/acprof:oso/9780199252695.001.0001} {\emph{{Efficiency and Complexity in Grammars}}}.
\newblock Oxford University Press.

\bibitem[{Heim and Kratzer(1998)}]{heim-kratzer-1998}
Irene Heim and Angelika Kratzer. 1998.
\newblock \href {https://www.wiley.com/en-us/Semantics+in+Generative+Grammar-p-9780631197133} {\emph{Semantics in Generative Grammar}}.
\newblock Wiley-Blackwell.

\bibitem[{Isono(2024)}]{isono-2024}
Shinnosuke Isono. 2024.
\newblock \href {https://doi.org/10.1016/j.cognition.2024.105766} {Category locality theory: A unified account of locality effects in sentence comprehension}.
\newblock \emph{Cognition}, 247:105766.

\bibitem[{Jaeger and Tily(2011)}]{jaeger-tily-2011}
T.~Florian Jaeger and Harry Tily. 2011.
\newblock \href {https://doi.org/10.1002/wcs.126} {On language ‘utility’: processing complexity and communicative efficiency}.
\newblock \emph{WIREs Cognitive Science}, 2(3):323--335.

\bibitem[{Kemp and Regier(2012)}]{kemp-regier-2012}
Charles Kemp and Terry Regier. 2012.
\newblock \href {https://doi.org/10.1126/science.1218811} {Kinship categories across languages reflect general communicative principles}.
\newblock \emph{Science}, 336(6084):1049--1054.

\bibitem[{Kemp et~al.(2018)Kemp, Xu, and Regier}]{kemp-etal-2018}
Charles Kemp, Yang Xu, and Terry Regier. 2018.
\newblock \href {https://doi.org/10.1146/annurev-linguistics-011817-045406} {Semantic typology and efficient communication}.
\newblock \emph{Annual Review of Linguistics}, 4(1):109--128.

\bibitem[{Kingma and Ba(2015)}]{kingma-ba-2015}
Diederik~P. Kingma and Jimmy Ba. 2015.
\newblock Adam: A method for stochastic optimization.
\newblock In \emph{Proceedings of the International Conference Learning Representations}, San Diego, CA, USA. Conference Track Proceedings.

\bibitem[{Kirby et~al.(2015)Kirby, Tamariz, Cornish, and Smith}]{kirby-etal-2015}
Simon Kirby, Monica Tamariz, Hannah Cornish, and Kenny Smith. 2015.
\newblock \href {https://doi.org/10.1016/j.cognition.2015.03.016} {Compression and communication in the cultural evolution of linguistic structure}.
\newblock \emph{Cognition}, 141:87--102.

\bibitem[{Kuncoro et~al.(2018)Kuncoro, Dyer, Hale, Yogatama, Clark, and Blunsom}]{kuncoro-etal-2018}
Adhiguna Kuncoro, Chris Dyer, John Hale, Dani Yogatama, Stephen Clark, and Phil Blunsom. 2018.
\newblock \href {https://doi.org/10.18653/v1/P18-1132} {{LSTM}s can learn syntax-sensitive dependencies well, but modeling structure makes them better}.
\newblock In \emph{Proceedings of the 56th Annual Meeting of the Association for Computational Linguistics (Volume 1: Long Papers)}, pages 1426--1436, Melbourne, Australia. Association for Computational Linguistics.

\bibitem[{Levy(2008)}]{levy-2008}
Roger Levy. 2008.
\newblock \href {https://doi.org/10.1016/j.cognition.2007.05.006} {Expectation-based syntactic comprehension}.
\newblock \emph{Cognition}, 106(3):1126--1177.

\bibitem[{Lewis and Vasishth(2005)}]{lewis-vasishth-2005}
Richard~L. Lewis and Shravan Vasishth. 2005.
\newblock \href {https://doi.org/10.1207/s15516709cog0000\_25} {An activation-based model of sentence processing as skilled memory retrieval}.
\newblock \emph{Cognitive Science}, 29(3):375--419.

\bibitem[{Lopopolo et~al.(2017)Lopopolo, Frank, {van den Bosch}, and Willems}]{lopopolo-etal-2017}
Alessandro Lopopolo, Stefan~L. Frank, Antal {van den Bosch}, and Roel~M. Willems. 2017.
\newblock \href {https://doi.org/10.1371/journal.pone.0177794} {Using stochastic language models ({SLM}) to map lexical, syntactic, and phonological information processing in the brain}.
\newblock \emph{PLoS ONE}, 12(5):e0177794.

\bibitem[{Mollica et~al.(2021)Mollica, Bacon, Zaslavsky, Xu, Regier, and Kemp}]{mollica-etal-2021}
Francis Mollica, Geoff Bacon, Noga Zaslavsky, Yang Xu, Terry Regier, and Charles Kemp. 2021.
\newblock \href {https://doi.org/10.1073/pnas.2025993118} {The forms and meanings of grammatical markers support efficient communication}.
\newblock \emph{Proceedings of the National Academy of Sciences}, 118(49):e2025993118.

\bibitem[{Montague(1970)}]{montague-1970}
Richard Montague. 1970.
\newblock \href {https://doi.org/10.1111/j.1755-2567.1970.tb00434.x} {Universal grammar}.
\newblock \emph{Theoria}, 36(3):373--398.

\bibitem[{Nivre et~al.(2020)Nivre, de~Marneffe, Ginter, Haji{\v{c}}, Manning, Pyysalo, Schuster, Tyers, and Zeman}]{nivre-etal-2020}
Joakim Nivre, Marie-Catherine de~Marneffe, Filip Ginter, Jan Haji{\v{c}}, Christopher~D. Manning, Sampo Pyysalo, Sebastian Schuster, Francis Tyers, and Daniel Zeman. 2020.
\newblock \href {https://aclanthology.org/2020.lrec-1.497} {{U}niversal {D}ependencies v2: An evergrowing multilingual treebank collection}.
\newblock In \emph{Proceedings of the Twelfth Language Resources and Evaluation Conference}, pages 4034--4043, Marseille, France. European Language Resources Association.

\bibitem[{Noji and Oseki(2021)}]{noji-oseki-2021}
Hiroshi Noji and Yohei Oseki. 2021.
\newblock \href {https://doi.org/10.18653/v1/2021.findings-acl.380} {Effective batching for recurrent neural network grammars}.
\newblock In \emph{Findings of the Association for Computational Linguistics: ACL-IJCNLP 2021}, pages 4340--4352, Online. Association for Computational Linguistics.

\bibitem[{Partee and Rooth(1983)}]{partee-rooth-1983}
Barbara Partee and Mats Rooth. 1983.
\newblock \href {https://doi.org/10.1515/9783110852820.361} {Generalized conjunction and type ambiguity}.
\newblock \emph{Formal semantics: The essential readings}, pages 334--356.

\bibitem[{Partee(1970)}]{partee-1970}
Barbara~Hall Partee. 1970.
\newblock \href {https://www.jstor.org/stable/25000447} {Negation, conjunction, and quantifiers: Syntax vs.\ semantics}.
\newblock \emph{Foundations of Language}, 6:153--165.

\bibitem[{Piantadosi et~al.(2011)Piantadosi, Tily, and Gibson}]{piantadosi-etal-2011}
Steven~T. Piantadosi, Harry Tily, and Edward Gibson. 2011.
\newblock \href {https://doi.org/10.1073/pnas.1012551108} {Word lengths are optimized for efficient communication}.
\newblock \emph{Proceedings of the National Academy of Sciences}, 108(9):3526--3529.

\bibitem[{Piantadosi et~al.(2012)Piantadosi, Tily, and Gibson}]{piantadosi-etal-2012}
Steven~T. Piantadosi, Harry Tily, and Edward Gibson. 2012.
\newblock \href {https://doi.org/10.1016/j.cognition.2011.10.004} {The communicative function of ambiguity in language}.
\newblock \emph{Cognition}, 122(3):280--291.

\bibitem[{Pimentel et~al.(2023)Pimentel, Meister, Wilcox, Mahowald, and Cotterell}]{pimentel-etal-2023}
Tiago Pimentel, Clara Meister, Ethan Wilcox, Kyle Mahowald, and Ryan Cotterell. 2023.
\newblock \href {https://doi.org/10.18653/v1/2023.emnlp-main.137} {Revisiting the optimality of word lengths}.
\newblock In \emph{Proceedings of the 2023 Conference on Empirical Methods in Natural Language Processing}, pages 2240--2255, Singapore. Association for Computational Linguistics.

\bibitem[{Regier et~al.(2015)Regier, Kemp, and Kay}]{regier-etal-2015}
Terry Regier, Charles Kemp, and Paul Kay. 2015.
\newblock \href {https://doi.org/10.1002/9781118346136.ch11} {Word meanings across languages support efficient communication}.
\newblock In \emph{The Handbook of Language Emergence}, chapter~11, pages 237--263. John Wiley \& Sons, Ltd.

\bibitem[{Resnik(1992)}]{resnik-1992}
Philip Resnik. 1992.
\newblock \href {https://aclanthology.org/C92-1032} {Left-corner parsing and psychological plausibility}.
\newblock In \emph{{COLING} 1992 Volume 1: The 14th {I}nternational {C}onference on {C}omputational {L}inguistics}.

\bibitem[{Ross(1967)}]{ross-1967}
John~R. Ross. 1967.
\newblock \emph{Constraints on variables in syntax}.
\newblock Ph.D. thesis, MIT.

\bibitem[{Shain et~al.(2020)Shain, Blank, {van Schijndel}, Schuler, and Fedorenko}]{shain-etal-2020}
Cory Shain, Idan~Asher Blank, Marten {van Schijndel}, William Schuler, and Evelina Fedorenko. 2020.
\newblock \href {https://doi.org/10.1016/j.neuropsychologia.2019.107307} {{fMRI} reveals language-specific predictive coding during naturalistic sentence comprehension}.
\newblock \emph{Neuropsychologia}, 138:107307.

\bibitem[{Shain et~al.(2024)Shain, Meister, Pimentel, Cotterell, and Levy}]{shain-etal-2024}
Cory Shain, Clara Meister, Tiago Pimentel, Ryan Cotterell, and Roger Levy. 2024.
\newblock \href {https://doi.org/10.1073/pnas.2307876121} {Large-scale evidence for logarithmic effects of word predictability on reading time}.
\newblock \emph{Proceedings of the National Academy of Sciences}, 121(10):e2307876121.

\bibitem[{Smith and Levy(2013)}]{smith-levy-2013}
Nathaniel~J. Smith and Roger Levy. 2013.
\newblock \href {https://doi.org/10.1016/j.cognition.2013.02.013} {The effect of word predictability on reading time is logarithmic}.
\newblock \emph{Cognition}, 128(3):302--319.

\bibitem[{Stanojevi\'{c} et~al.(2023)Stanojevi\'{c}, Brennan, Dunagan, Steedman, and Hale}]{stanojevic-etal-2023}
Milo\v{s} Stanojevi\'{c}, Jonathan~R. Brennan, Donald Dunagan, Mark Steedman, and John~T. Hale. 2023.
\newblock \href {https://doi.org/10.1111/cogs.13312} {Modeling structure-building in the brain with {CCG} parsing and {L}arge {L}anguage {M}odels}.
\newblock \emph{Cognitive Science}, 47(7):e13312.

\bibitem[{Steinert-Threlkeld(2021)}]{steinert-threlkeld-2021}
Shane Steinert-Threlkeld. 2021.
\newblock \href {https://doi.org/10.3390/e23101335} {Quantifiers in natural language: Efficient communication and degrees of semantic universals}.
\newblock \emph{Entropy}, 23(10).

\bibitem[{Stern et~al.(2017)Stern, Fried, and Klein}]{stern-etal-2017}
Mitchell Stern, Daniel Fried, and Dan Klein. 2017.
\newblock \href {https://doi.org/10.18653/v1/D17-1178} {Effective inference for generative neural parsing}.
\newblock In \emph{Proceedings of the 2017 Conference on Empirical Methods in Natural Language Processing}, pages 1695--1700, Copenhagen, Denmark. Association for Computational Linguistics.

\bibitem[{Sturt and Lombardo(2005)}]{sturt-lombardo-2005}
Patrick Sturt and Vincenzo Lombardo. 2005.
\newblock \href {https://doi.org/10.1207/s15516709cog0000\_8} {Processing coordinated structures: Incrementality and connectedness}.
\newblock \emph{Cognitive Science}, 29(2):291--305.

\bibitem[{Trott and Bergen(2022)}]{trott-bergen-2022}
Sean Trott and Benjamin Bergen. 2022.
\newblock \href {https://doi.org/10.1016/j.cognition.2022.105094} {Languages are efficient, but for whom?}
\newblock \emph{Cognition}, 225:105094.

\bibitem[{Uegaki(2022)}]{uegaki-2022}
Wataru Uegaki. 2022.
\newblock \href {https://doi.org/10.1162/ling_a_00461} {{The Informativeness/Complexity Trade-Off in the Domain of Boolean Connectives}}.
\newblock \emph{Linguistic Inquiry}, pages 1--23.

\bibitem[{{van de Pol} et~al.(2023){van de Pol}, Lodder, {van Maanen}, Steinert-Threlkeld, and Szymanik}]{van-de-pol-etal-2023}
Iris {van de Pol}, Paul Lodder, Leendert {van Maanen}, Shane Steinert-Threlkeld, and Jakub Szymanik. 2023.
\newblock \href {https://doi.org/10.1016/j.cognition.2022.105150} {Quantifiers satisfying semantic universals have shorter minimal description length}.
\newblock \emph{Cognition}, 232:105150.

\bibitem[{{van Schijndel} et~al.(2013){van Schijndel}, Exley, and Schuler}]{van-schijndel-etal-2013}
Marten {van Schijndel}, Andy Exley, and William Schuler. 2013.
\newblock \href {https://doi.org/10.1111/tops.12034} {A model of language processing as hierarchic sequential prediction}.
\newblock \emph{Topics in Cognitive Science}, 5(3):522--540.

\bibitem[{White and Cotterell(2021)}]{white-cotterell-2021}
Jennifer~C. White and Ryan Cotterell. 2021.
\newblock \href {https://doi.org/10.18653/v1/2021.acl-long.38} {{Examining the Inductive Bias of Neural Language Models with Artificial Languages}}.
\newblock In \emph{Proceedings of the 59th Annual Meeting of the Association for Computational Linguistics and the 11th International Joint Conference on Natural Language Processing (Volume 1: Long Papers)}, pages 454--463, Online. Association for Computational Linguistics.

\bibitem[{Zaslavsky et~al.(2018)Zaslavsky, Kemp, Regier, and Tishby}]{zaslavsky-etal-2018}
Noga Zaslavsky, Charles Kemp, Terry Regier, and Naftali Tishby. 2018.
\newblock \href {https://doi.org/10.1073/pnas.1800521115} {Efficient compression in color naming and its evolution}.
\newblock \emph{Proceedings of the National Academy of Sciences}, 115(31):7937--7942.

\bibitem[{Zipf(1949)}]{zipf-1949}
George~K. Zipf. 1949.
\newblock \emph{Human behavior and the principle of least effort}.
\newblock Addison-Wesley.

\end{thebibliography}



\end{document}